\documentclass[final,3p,times]{elsarticle}
\usepackage{amssymb}
\usepackage{graphicx}
\usepackage{subcaption}
\usepackage{times}
\usepackage{xcolor}
\usepackage{soul}
\usepackage[utf8]{inputenc}
\usepackage{graphicx, wrapfig}
\usepackage{booktabs}
\usepackage{amsthm}
\newcommand{\ZN}{\texttt{ZN} }
\newcommand{\TE}{\texttt{Text-Encoder} }
\newcommand{\ZC}{\texttt{Zooming-Controller} }

\journal{Neural Computing}

\begin{document}

\begin{frontmatter}



\title{Zooming Network}


\author{YukunYan$^1$, 
Daqi Zheng$^2$, 
Zhengdong Lu$^2$, 
Sen Song$^1$
\\ 
$^1$ Tsinghua University \\
$^2$ DeeplyCurious.ai\\
{yanyk13, songsen}@mails.tsinghua.edu.cn,
{da, luz}@deeplycurious.ai}

\address{}

\begin{abstract}
Structural information is important in natural language understanding. Although some current neural net-based models have a limited ability to take local syntactic information, they fail to use high-level and large-scale structures of documents. This information is valuable for text understanding since it contains the author's strategy to express information, in building an effective representation and forming appropriate output modes. We propose a neural net-based model, \texttt{Zooming Network}, capable of representing and leveraging text structure of long document and developing its own analyzing rhythm to extract critical information. Generally, \ZN consists of an encoding neural net(\TE) that can build a hierarchical representation of a document, and an interpreting neural model that can read the information at multi-levels and issuing labeling actions through a policy-net (\ZC). Our model is trained with a hybrid paradigm of supervised learning (distinguishing right and wrong decision) and reinforcement learning (determining the goodness among multiple right paths).  We applied the proposed model to long text sequence labeling tasks, with performance  exceeding baseline model (biLSTM-crf) by 10 F1-measure.
\end{abstract}


\end{frontmatter}


\section{Introduction}
Although theoretically, a vanilla DNN model such as a multi-layer perceptron with non-linear activation-functions can fit any functions, it is impossible to gather enough labeled data in most situations. Exquisitely designed neural-based models introduce some architecture bias and operation bias to reduce the dependency on data amount and achieve excellent performances on different problems.  One typical instance is Convolutional Neural Networks, which uses shared convolution kernels and max-pooling operations to extract features. Such architecture perfectly suits the attributes of digital images and achieves great success in computer vision tasks. However, there is no general architecture available in natural language processing.

Natural languages have several characteristics as follows: 1) A paragraph of text is a sequence of letters or characters, which orderly express information. 2) A higher-level language unit consists of an arbitrary number of lower-level language units. Correspondingly, local semantic feature together makes up a more complex expression. 3) Unlike images, text information is more discretized and symbolized. We aim to design a neural-based network, which has better natural language processing performance leveraging such characteristics.

In this paper, we focus on a classic task in natural language processing, name entity recognition(NER). NER is a typical sequence to sequence decision progress, in which, generally, given a paragraph, a model finds out several keys fragments such as names, locations or organizations by indicating them via special labels. Several models are introduced recently such as combinations of convolutional layers, gated recurrent layers, and conditional random field. However, these models are still some distance away from practical applications in the real world for following reasons: 1)  Current neural based models become powerless establishing representations when a paragraph has a complex structure, which always leads to more tanglesome dependencies. 2) They lack the ability of dynamically utilizing different levels of details to maintain a more effective memory, which plays an important role in human reading and understanding. 3) The output mode, one single label per time-step, is inefficient and not necessary, which takes much time when processing a long paragraph.

To conquer such weakness of current networks and efficiently utilizing language characteristics, we propose a novel framework, Zooming Network(\ZN).  \ZN explicitly uses text structure information of language units segmentation and affiliation to establish multi-level representations of a text during encoding section.  Then, it reads these representations in order via a stochastic policy and outputs an arbitrary number of labels in the meantime. The structure of \ZN is roughly an encoder-controller-reasoner framework, as illustrated in Figure1. The first part called Hierarchical Encoder maps a document into a hierarchical representation, where each language unit has a corresponding distributed memory slot. The second part named Skipping-Controller follows a stochastic policy to read part of earlier established memory and output a predicted sequence. Meanwhile, as the third part of \ZN,  the Symbolic Reasoner provides some symbolic clues to SC to help make better choices. We elaborate details in next several sections.

\begin{figure}[h]
	\centering
	\includegraphics[width=0.5\textwidth]{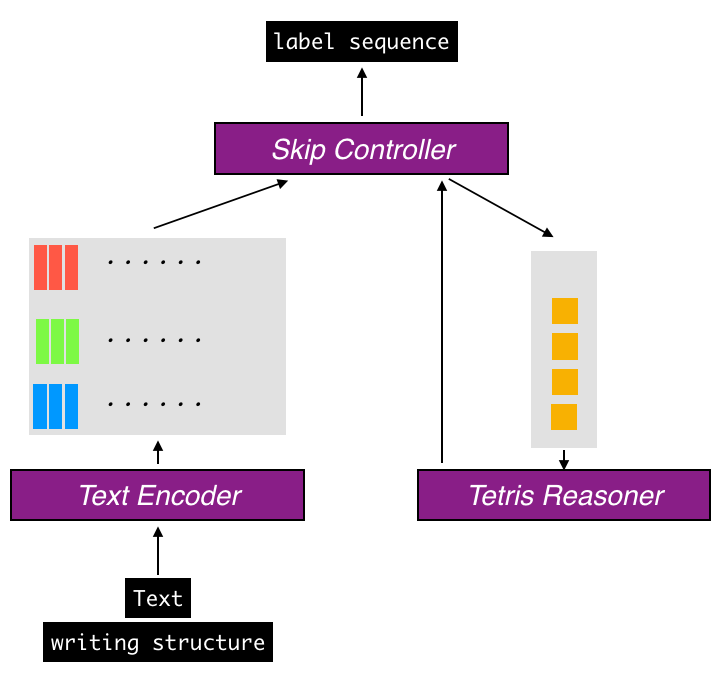}
	\caption{scale-free Identification Network: Skip Controller}
	\label{fig:sfIN_SC}
\end{figure}

\section{Related Work}
The idea of using multi-resolution features has been explored in various ways. Some early inspiring attempts date back to the 90's, such like \cite{el1996hierarchical}, \cite{lin1998learning} and \cite{schmidhuber1991neural}. In such works, models have multiple layers of RNNs in a decreasing order of update frequency. The clockwork RNN (CW-RNN) \cite{koutnik2014clockwork} is a more recent model extended from hierarchical RNN \cite{el1996hierarchical} and NARX RNN \cite{lin1996learning}, in which the hidden layer is partitioned into separate modules, each processing inputs at its own temporal granularity, making computations only at its prescribed clock rate. However, the CW-RNN explicitly assigning hard time-scale to update its hidden state, which is counterintuitive for that language structures have various lengths. The biscale RNNs \cite{chung2015gated} try to solve this problem by introducing \textsl{layer-wise} timescales by having additional gating units. However, it relies on the soft gating mechanism like LSTMs, which is computationally expensive and limited to the capability of memorizing long-term memory. In the contrast, \ZN explicitly uses hierarchical text structural information to obtain a multi-resolution representation with increasing levels of abstraction.

Other models capable of using explicit boundary structure are proposed recently. The hierarchical RNN in \cite{ling2015character} uses word boundary via tokenization for machine translation by modeling character-level and word-level representation using two individual RNN layers. A similar model is used in \cite{sordoni2015hierarchical} to model dialogue utterances. However, these models use relatively low-level boundary information in short text, while \ZN focus on high-level text structure such as sentence and paragraph in sequence labeling tasks of long texts.

Different from all of the previous models, \ZN not only leverages text structure information in representation establishment but also in forming a new output mode. As far as we know, \ZN is the first model capable of labeling a sequence with multi-level actions. In this way, structure information is used more flexible and direct to model long-term dependencies.
The paper 'Hierarchical Attention Networks for Document Classification' has the following important differences as sFIN : 1) HAN encodes a document into a single vector, which is used in classification, while \ZN keeps information in all three granularities then makes a series of decisions. 2)  sfIN also utilizes structure information in decision making progress. 3) HAN aims to find some important words/sentences via hierarchical attention mechanism to better classify a document while sfIN use structure information to find a better way to update the state of the controller.  We are sorry that we didn't make it clear that our tagging is dynamical:  it can generate unbounded tags in [0,1,2,..] which indicates the serial number of a target fragment. Corresponding to BIO annotation,  "none crucial", "current fragment", "new fragment"respectively stand for 'O', 'I', 'B'. 

\section{Zooming Network}
We take a NER task as a decision progress, which can be roughly divided into two steps: information representation and decision making. A paragraph of text is essentially a sequence of basic language units, letters or words and an arbitrary number of them make up a higher-level language unit, like a sentence or a paragraph. Considering this, we establish the representation of a text from front to back and from bottom to top. In general, the representation of basic unit (like word-embedding) is built first, and the units in next level are encoded using such basic information. Inspired by multi-layer convolutional neural networks, we use the max-pooling operation to calculate the abstract of lower-level information as the representation of current-level ones. After hierarchical representation established by \TE, \ZC takes controller of label selection. It reads from front to back while simultaneously select an arbitrary number of labels and decide a proper \textsl{read path}. Details are illustrated in several following subsections.
  
\subsection{Text Encoder: Establishment of Hierarchical Representation}

\begin{figure}[h]
	\centering
	\includegraphics[width=1\textwidth]{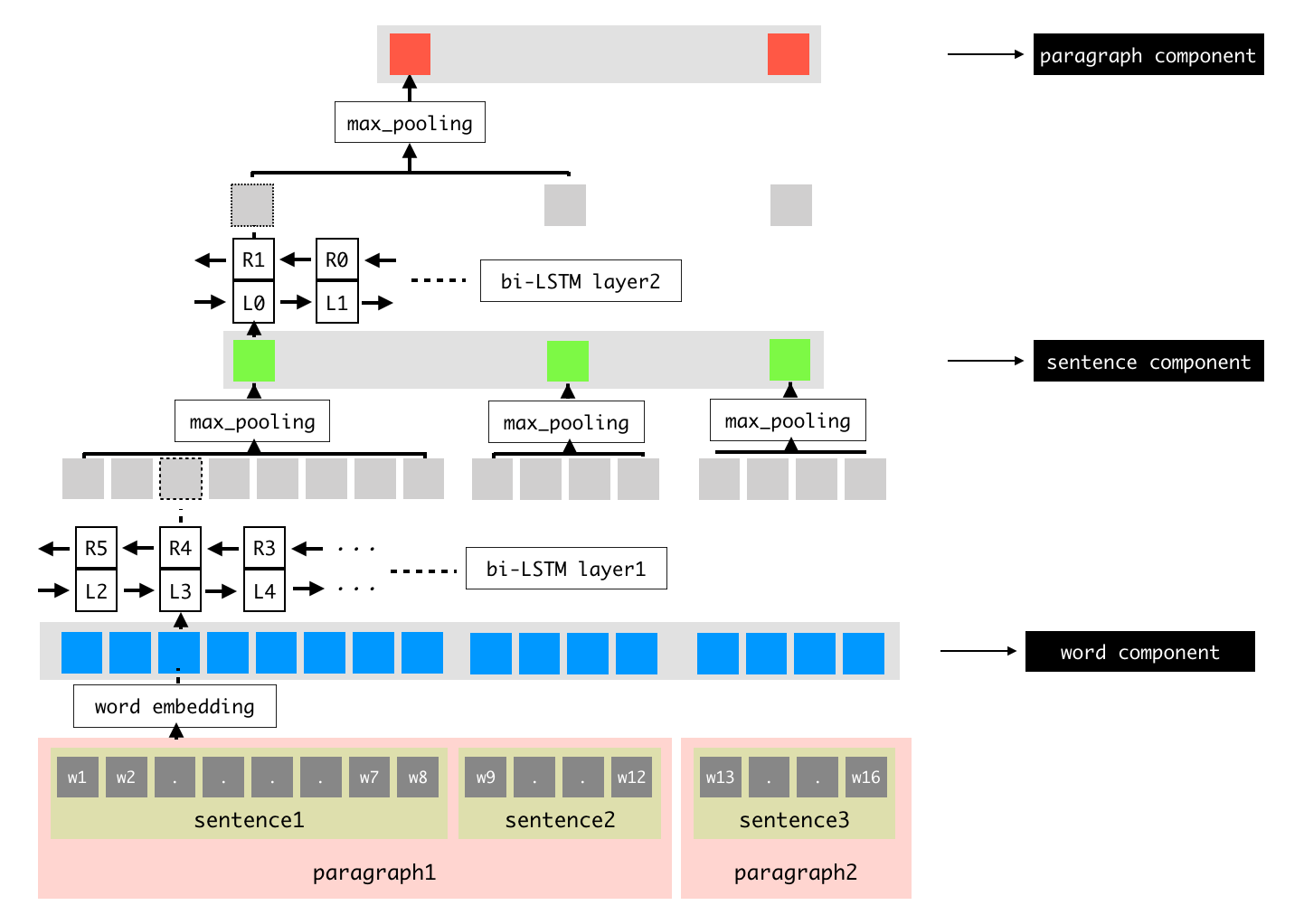}
	\caption{Zooming Network: Text-Encoder}
	\label{fig:te}
\end{figure}

\begin{figure}[h]
	\centering
	\includegraphics[width=0.6\textwidth]{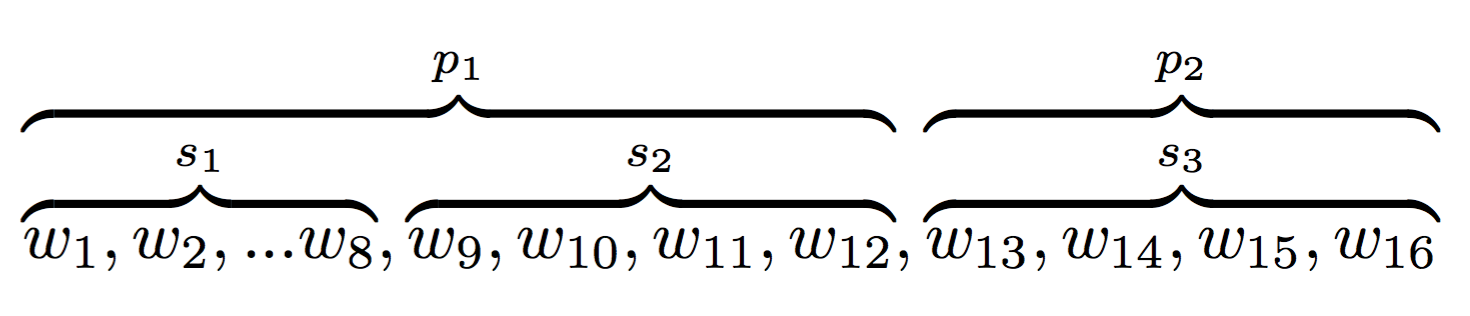}
	\caption{An example text}
	\label{fig:example}
\end{figure}

To begin with, we provide an example of how Zooming Network generates a hierarchical memory slot with its first part and then generates a label sequence with its second part. In Figure \ref{fig:example}, we depict a long text with two paragraphs, three sentences and an arbitrary number of words, where the affiliation is shown by the brackets. As illustrated in Figure \ref{fig:te}, the first part of ZN, Text Encoder has three layers,  word-embedding layer and two biLSTM layers. The first biLSTM layer receives words as inputs and generated sentence-level representation, and the second one takes word-level representations as input and yield paragraph-level representations. These three components compose the hierarchical memory, as shown in Figure X. After such encoding progress, every language unit has a corresponding real-value vector as a distributed representation. From this example we can see the advantages of Zooming Network: 1) The update times of each recurrent unit is remarkably less then that of a single recurrent network, which make it much easier modeling long-term dependencies and temporal information. 2) The upper level representation is summarized from the lower level information with explicit boundary information which make it capable of modeling spatial information similar with convolutional neural networks. 

The word-level component of the hierarchical memory is first generated by the word-embedding layer, which is initialized by word2vec, and then fine tuned together with the other parts of the whole model.
$$v_{w_i} = f_{embed}(w_i) \eqno{1}$$
$$R_w = [v_{w_1}, v_{w_2} ... v_{w_n}] \eqno{2}$$

The first bi-LSRM layer takes the word vectors from each sentence as input and outputs a hidden state vector at each time step. Then we use the max-pooling operation to gain a fixed length sentence vector. We take all the sentence vectors as the sentence-level component of the hierarchical memory.

$$v_{s_j} = max(f_{bi-LSTM1}(v_{w_i}, v_{w_{i+1}} ... v_{w_n})),  w_{i}, w_{i+1} ... w_{n} \in s_j \eqno{3} $$
$$R_s = [v_{s_1}, v_{s_2} ... v_{s_m}] \eqno{4}$$

The paragraph-level component is established in a similar way. We use the second bi-LSTM layer and the max-pooling operation to encode sentence-level information into paragraph-level representation.

$$v_{p_j} = max(f_{bi-LSTM2}(v_{s_j}, v_{s_{j+1}} ... v_{s_m})),  s_{j}, s_{j+1} ... s_{m} \in p_k \eqno{5} $$
$$R_p = [v_{p_1}, v_{p_2}, ... v_{p_o}] \eqno{6}$$ 

We concatenate $R_w$, $R_s$, $R_p$ to gain the hierarchical memory $M=[R_w, R_s, R_p]$.

\subsection{Zooming Controller: Zooming while Generating Labels}
To make the description more readable, we first introduce some concepts and basic configuration before elaborating the \ZC: 
\begin{itemize}
	\item \textbf{read-head}: \ZN has three read-heads, each of which reads one vector in the hierarchical representation at a specific level at current time step.
	\item \textbf{location vector}: The location vector, $v_l\in Z^3$ declares the location of the \textsl{read-heads}. For example, $v_l = [1, 1, 1]$ means the \textsl{read-heads} are pointing at the first word, the first sentence, and the first paragraph.
	\item \textbf{actions}: Instead of directly choosing one simple label at each time step, \ZC chooses and executes, which has a specific level and a specific type an action to generate labels. 
	\item \textbf{action level}: Action level indicates the object of corresponding action is at which level, with three available options: word-level, sentence-level, and paragraph-level.
	\item \textbf{action type}: BIO is a classic label standard in a name-entity-recognition task. Since an action is can be at a sentence-level or paragraph level, we propose some corresponding action type to the BIO standard. 1)'O' indicates current fragment is outside of any mention of a named entity. 2)'B' indicates current fragment is the beginning of a mentioned named entity. 3)'I' indicates current fragment is inside of a mention of a named entity. By current fragment we mean the object language unit of current action.
	\item \textbf{previous action vector}: $v_p$ is an one-hot vector of length nine, indicating the action executed at last time step. 
	\item \textbf{action execution}: The execution progress of an action is generating a label sub sequence based on the type and level of an action. For example, suppose the sentence-level read-head is pointing a sentence with length 5 when \ZC takes an sentence-level 'O' action, the corresponding label sequence generated at current time step should be $['O', 'O', 'O', 'O', 'O']$
	\item \textbf{skipping rule}: After the execution of an action, the three read-heads should skip to the next location according to following rules: 1) If it is an word-level action, the word-level read-head first turn to the next word, and the other two read-heads then change/keep their position based on the belonging situation of the very word. 2) If it is an sentence-level action, the sentence-level first skip to the next sentence, then the word-level read-head turns to the first word of such sentence. And the paragraph-level read-head changes/keeps its location based on the belonging situation of the very sentence. 3) If it is a paragraph-level action, the paragraph-level read-head first turn to the next paragraph, and then the other two read-heads turn to the next word and the next sentence.
	\item \textbf{processing path}: We take the location history of the word-level read-head as the processing path.
	\item \textbf{action history}: We take the action execution history as action history.
\end{itemize} 

\textsl{How does the Zooming Controller generate the label sequence by executing a series of  actions?} We use the example in Figure X agin to give a rough description of this special decoding progress. After the encoding procedure,  the model start to process the whole text from the very beginning. The representation of $w_1$, $s_1$, and $p_1$ and the previous action vector, $v_p$, which initialized as an all-zero vector are concatenated as the input vector. The RNN layer of Zooming Controller then takes the input vector and update its hidden state, which is used to predict an action.  Suppose an ‘B’-sentence-level action is chosen, the corresponding label, $[B, I, I, I, I, I, I, I]$ is generated as the label of current sentence. Following the skipping rule, After the execution of the action, the location vector updates to $[9, 2, 1]$, and the previous action vector updates to $[0, 0, 0, 1, 0, 0, 0, 0, 0]$.  After this, predictions, executions, and updates are looped until the entire text is processed. In this case, another four 'O'-word-level actions and one 'B'-paragraph-level action are executed, and the corresponding predicted labels are concatenated together as the result. We abstract the whole process into the following parts

\begin{figure}[h]
	\centering
	\includegraphics[width=0.6\textwidth]{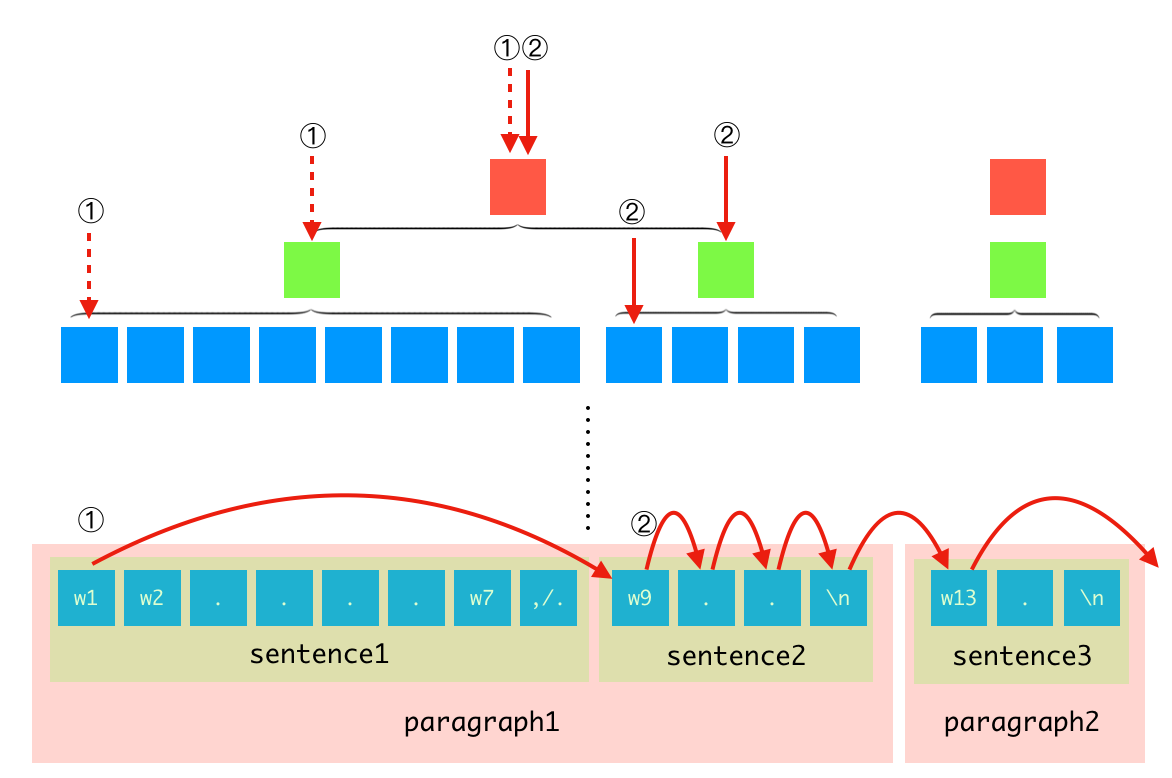}
	\caption{The processing of the example}
	\label{fig:contollrt_example}
\end{figure}

\begin{itemize}
	\item \textbf{Initialization}: After the hierarchical representation established, we first initialize the location vector $v_{l_0}$ as $[1, 1, 1]$, the previous action vector $v_{p_0}$ as $[0, 0, 0, 0, 0, 0, 0, 0, 0]$, and the predicted label as an empty list.
		$$v_{l_0} = [1, 1, 1] \eqno{7}$$
		$$v_{p_0} = [0, 0, 0, 0, 0, 0, 0, 0, 0] \eqno{8}$$
		$$l_0 = [] \eqno{9}$$
	\item \textbf{Prediction}: Guided by the location vector, each read-head reads a specific part of the hierarchical representation (a vector) in corresponding position. All three level vectors and the previous action vector are concatenated as current input vector $v_{i_t}$, which is then used to update the hidden state of RNN layer and choose the action at current time step through a fully connected layer and a softmax operation.
		$$v_{input}=[R_w[v_{l_t}[1]], R_s[v_{l_t}[2]], R_p[v_{l_t}[3]], v_{p_t}] \eqno{10}$$
		$$h_t = f_{RNN}(v_i, h_{t-1}) \eqno{11}$$
		$$a_t = argmax(softmax(f_c(h_t))) \eqno{12}$$
	\item \textbf{Execution}: After chosen an action, a corresponding label sequence is generated following the \textbf{action execution} rule, then appended to the end of the predicted label. 
		$$l_t = f_{execute}(a_t) \eqno{13}$$
		$$l_{p_t} = [l_{p_{t-1}}, l_t] \eqno{14}$$
	\item \textbf{Update}: Following the skip rule, the three read-heads move to the next positions. Meanwhile, the location vector $v_l$ and the previous action vector $v_p$ are updated correspondingly.
		$$v_{l_{t+1}} = f_{skipping}(v_{l_{t}}, a_t) \eqno{15}$$
		$$v_{p_{t+1}} = one-hot(a_t) \eqno{16}$$
	\item \textbf{Loop}: The \textbf{Prediction, Execution and Update} section are looped until the whole text is processed.
\end{itemize}

\begin{figure}[h]
	\centering
	\includegraphics[width=0.8\textwidth]{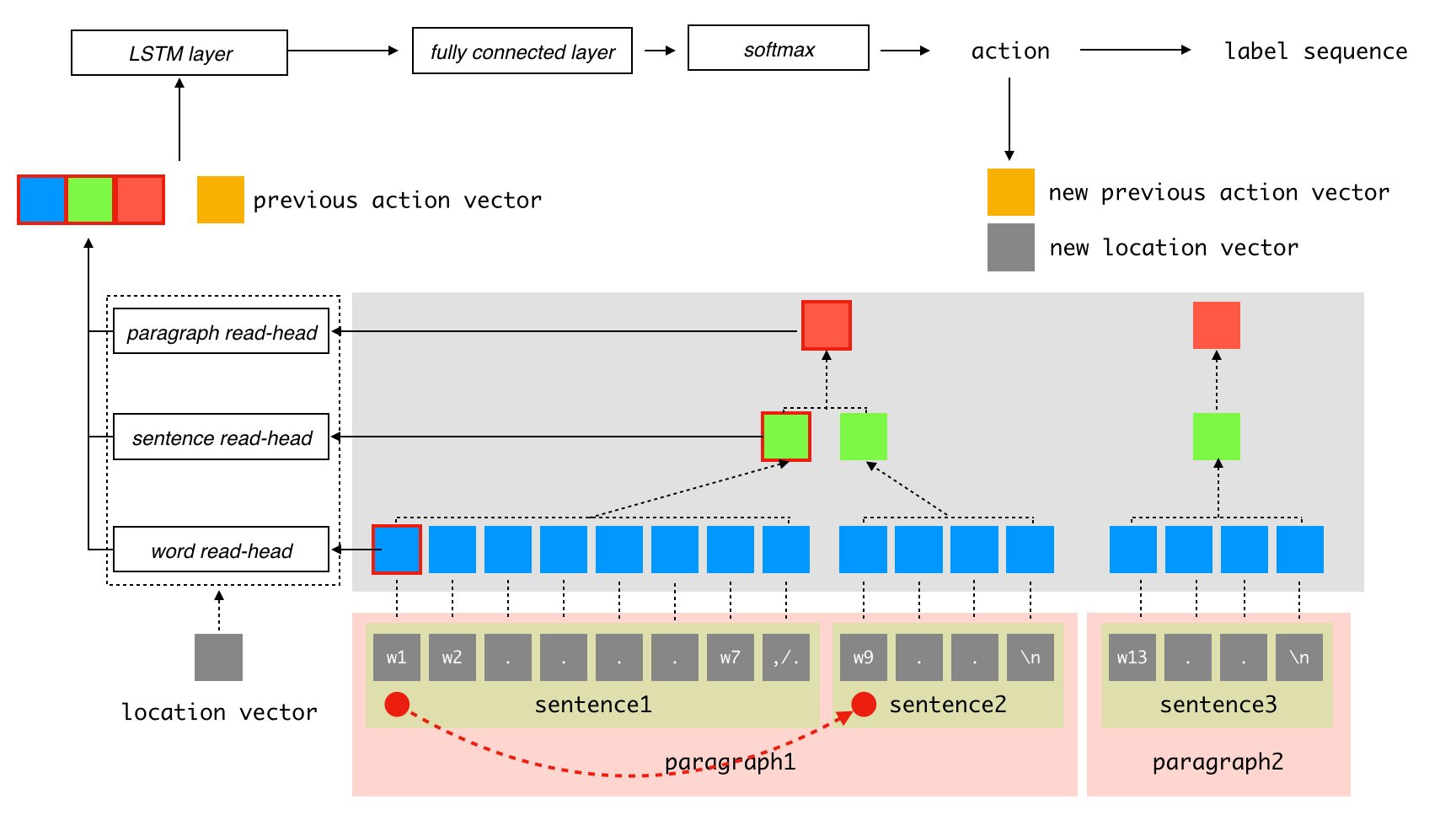}
	\caption{Zooming Controller}
	\label{fig:zooming_controller}
\end{figure}

The whole controller processing procedure is illustrated in Figure \ref{fig:zooming_controller}. As we can see from above description, \ZC reads multi-level information simultaneously, which gives the model three advantages: 1) a flexible resource allocation (e.g., The units that build higher-level representation focuses on modeling long-term dependencies, while lower-level representation related units focus on modeling local features). 2) efficiency delivering long-term dependencies with fewer updates of controller hidden state when executing a sentence/paragraph level action, which mitigates the vanishing gradient problem. 3) operation bias of a whole language unit, which avoids false gaps inside a large-scale important information fragment. 
\section{Learning}
Zooming Network is designed to 'zoom out' when the current segment is not highly correlated with the target information and 'zoom in' when find current information is significantly related to the target information. In other words, we hope that \ZN has an efficient processing path while maintain a good accuracy. We utilize supervised learning to make sure the model with a high accuracy and reinforcement learning to drive the model to use more high-level actions.  
\subsection{Supervised Learning}
As a sequence-to-sequence model, the optimization goal of \ZN is minimizing the distance between the ground truth distribution of label sequences $P^*(Y|X)$ and predicted distribution $P(Y|X)$ condition on the input sequence $X$. With the special design of \ZC, instead of directly outputs a label sequence, \ZN generates a sequence of actions, which makes it impossible to forthright leverage the ground truth label. In order to apply supervised learning, we convert the label into action supervision via the verification of whether the predicted label is correct after the execution of an action at each time step. As we can see, there can be more than one correct action at a specific time step. Take the text in Figure \ref{fig:example} as an example, at the very beginning, the 'B'-word-level action and 'B'-sentence-level action both can produce correct labels. We make all the correct actions into a set $A_t^*$. Since there is no way for us to know which action in $A_t^*$ is the best one (it involves the update path of the \ZC), we turn to maximize the sum possibility of $A_t^*$. Here we employ a modified cross entropy as the loss function as shown in equation (17).
$$L_\theta = -\frac{1}{T}\sum_{t=1}^T(\sum_{a_i\notin A_t^*}log(1-y_{a_i}) + log(\sum_{a_j \in A_t^*}y_{a_j})) \eqno{17}$$ 
%\begin{equation}
%l_{a_t}
%\left\{
%\begin{array}
%{r@{\quad\quad}l}
%1 & f_{execution}(a_t) = l_g[v_{l_t}[0]:v_{l_{t+1}}[0]]\\
%0  & else
%\end{array}
%\right.
%\end{equation}

\subsection{Path Reward}
As mentioned above, the processing procedure can be very versatile with multiple correct actions available at each time step. Experientially, a better path means skipping the irrelevant sections quickly, zooming into the region of interest and zooming out when there comes an non-important section again, which is what it is like for a human reader. However, it is not only the efficiency the processing path influences but also the state updating strategy. In another word, it is impossible to explicitly tell which region of the text can indicate its context is the important information, even for a human. Hence, we adopt a relatively conservative training strategy. We use an reinforcement learning method(policy gradient) to help tuning the whole network. The processing of a single text is regarded as an episode, at the end of which the model receives a reward indicating how well is the current processing path as shown in equation(18). 
$$r=L_\theta * (- \frac{N_{a_w}}{N_{a_w} + N_{a_s} + N_{a_p}}) \eqno{18}$$
where the $L_\theta$ is current loss and $N_{a_w}, N_{a_s}, N_{a_p}$ separately indicates the action number of each level actions. As we can see, the model is encouraged to use more high level actions to get a larger reward, or less $L_\theta$. Moreover, such configuration can guarantee that once the model achieves small enough cost, the processing path turn to be converged together with the predicted label sequences.
\subsection{Jointly Learning and Sampling Strategy}
\ZN is trained both using supervised learning and reinforcement learning, both involving gradient descent method. Intuitively, we leave the accuracy problem to supervised learning part, and use reinforcement learning to find a better processing path among hundreds of correct paths. Here, we introduce an hyper-parameter $\lambda$ to balance the supervised learning and reinforcement learning by giving a weighted sum gradient changes in equation (19).
$$J_\theta=-\sum_{t=1}^Tr_t * log(\pi_\theta(a_t, s_t))$$
$$L_\theta = -\frac{1}{T}\sum_{t=1}^T(\sum_{a_i\notin A_t^*}log(1-y_{a_i}) + log(\sum_{a_j \in A_t^*}y_{a_j}))$$
$$min(L_\theta - \lambda * J_\theta)$$
Since \ZC is essentially a sequence model with feedback inputs(we input the previous action vector into the controller module), we execute one of the correct actions at each time step during the training procedure following an on-policy method. In specific, we first calculate the probability of each correct action in the set $A_t^*$, and execute a proportional random selection from the very set. In this way, we guarantee that the model always follow a correct processing path during training and given the right feedback information(the previous action vector $v_p$).
\section{Experiments}

\subsection{TaskI:Event Extraction in Court Documents}
In this Task, we deal with some legal documents about some fact description, which is the detail information of a stealing case. Such samples shares following characters: 1) The text can be divided into several semantic sections. More specific, criminal case judgment contains the basic information of the criminal, the investigation procedure, the ascertained facts, the legal basis, the judgment conclusion and the specific judgments. 2) The information of interest only appears in a specific section. The detail information of stealing events are always placed in the asertained facts section. An example of the data is shown in Figure [].
Since the extraction of region of interest is a sequantial labeling task, we take one of the state of the art named entity recognition model, bidirectional Long short-term momery and conditional random field (biLSTM-crf) as the baseline model. We use the word-level accuracy and fragment-level f1 as the evaluation standard. The results are shown in Figure[]. To illustrate the processing efficiencies, we introduce a contant named  wlar(word level action ratio). A model with lower wlar means it has better processing efficiency.

\begin{figure}[h]
	\centering
	\includegraphics[width=0.5\textwidth]{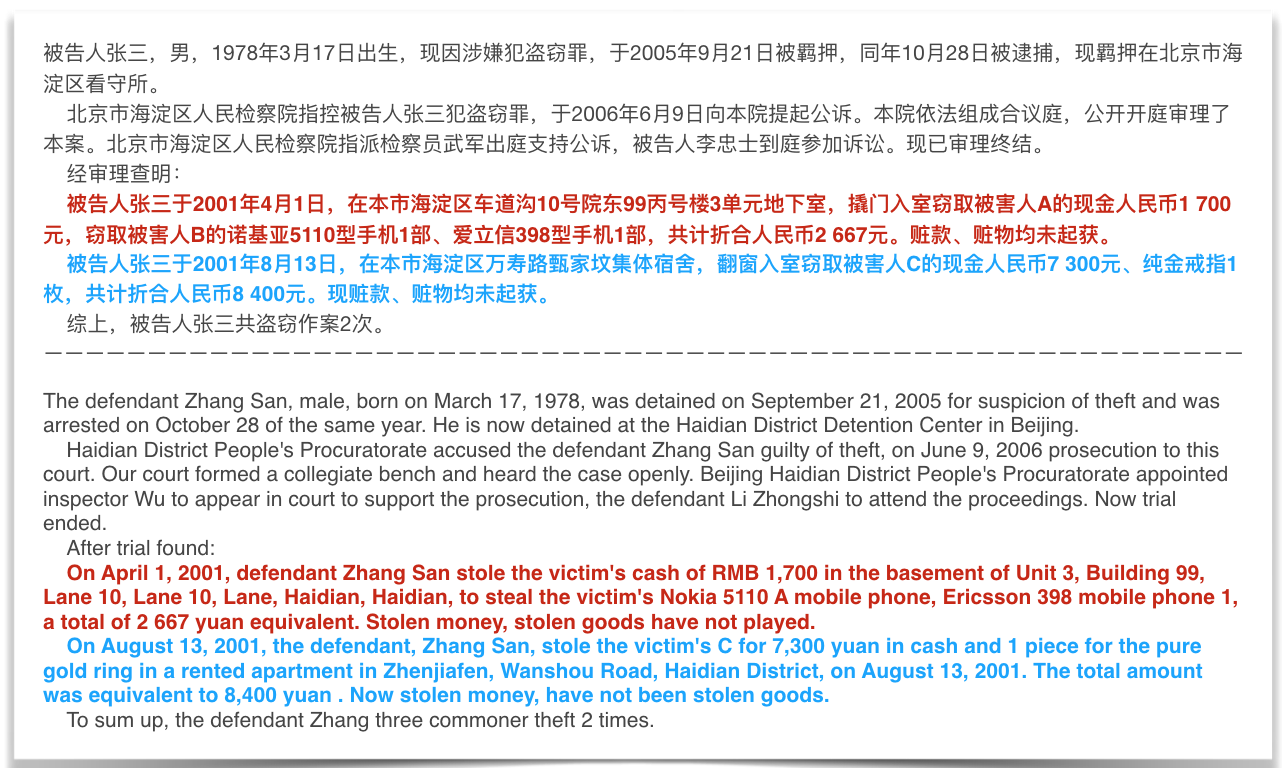}
	\caption{Part of an instance in TcJD: red part and blue part respectively indicts one incident}
	\label{fig:TcJD}
\end{figure}

We build this dataset including 3944 judgment documents about theft cases in Chinese. In such documents, the description of the court can be broadly divided into five parts: \textsl{basic information}, \textsl{the proposals of procuratorate}, \textsl{ascertain of facts}, \textsl{court's conclusion} and \textsl{judgment}. The identification task is to recognize theft incidents in both \textsl{the proposals of procuratorate} and \textsl{ascertain of facts} one by one, which is labeled at word-level. An instance is shown in Figure \ref{fig:TcJD} 

On average, a sample of \textsl{TcJD} has 1089.94 words, and 207.26 words of incident description.
\subsection{TaskII:Focus Extraction in Court Documents}
To give a step further test of \ZN, we design the second extraction task, the extraction of argument focuses in an intellectual property dispute cases. This dataset is built with 2180 judgment documents about patten infringement written in Chinese. Similar to \textsl{TcJD}, the documents can be roughly divided into several sections as well. The task is to identify focuses of dispute between the plaintiff and defendant.The focuses of the dispute is more varied than the description of the incident in \textsl{JcJD}, and its position is more uncertain. Two fragments of examples are shown below in Figure \ref{fig:PiJD}

\begin{figure}[h]
	\centering
	\includegraphics[width=0.5\textwidth]{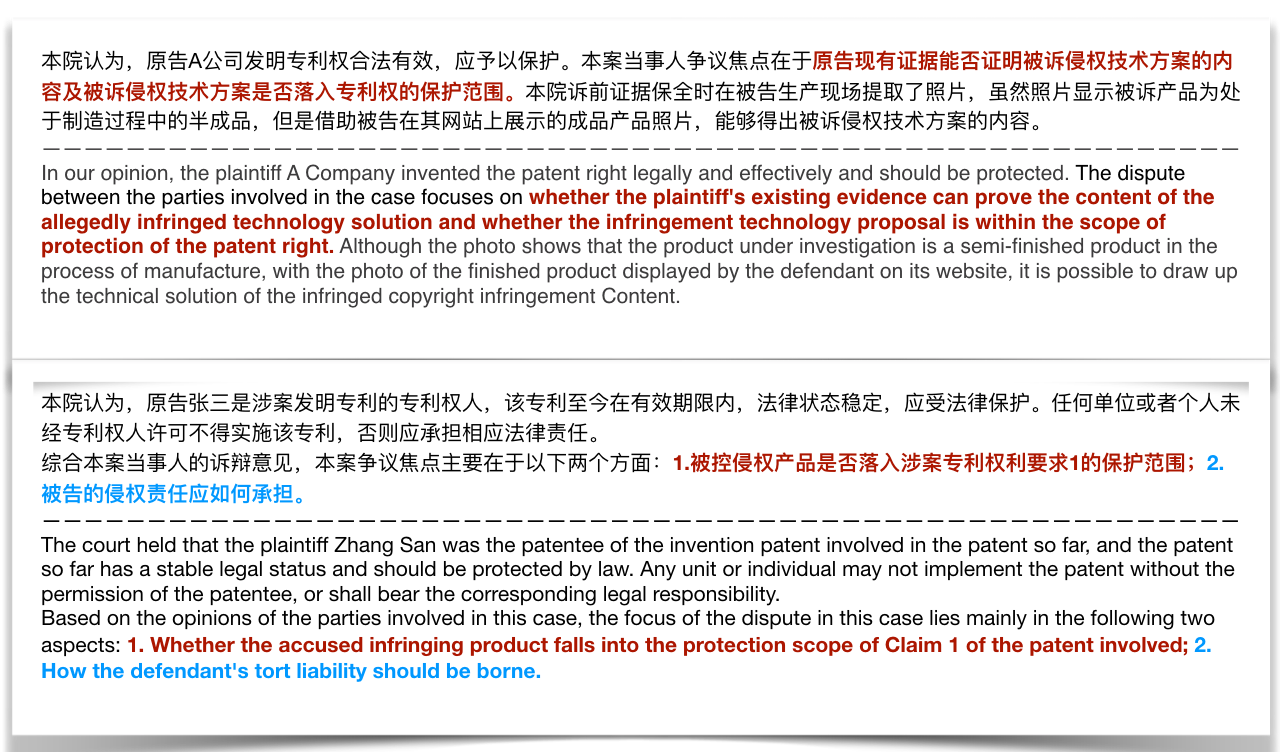}
	\caption{Two fragments of instances in PiJD: red part and blue part respectively indicts one focus of dispute}
	\label{fig:PiJD}
\end{figure}

On average, a sample of \textsl{PiJD} has 6133.61 words, and 82.11 words of description of focus.

\subsection{Evaluation Methods}
We use entity(or fragment)-level precision, recall and F1 measure to evaluate the performance of the proposed model. Meanwhile, we use \textsl{word-level actions ratio(wlar)} to evaluate the processing efficiency of our model, which is calculated as follows:
$$wlar = \frac{n_{a_w}}{n_w + n_s + n_p}$$
Lower wlar value infers that the model takes fewer actions in processing, which means more efficient.

\subsection{Implement Details}
We use \texttt{biLSTM-crf}, which is one of the state-of-the-art sequence labeling models, proposed in \cite{huang2015bidirectional}, as a baseline model. We set both \texttt{biLSTM-crf} and \ZN with similar number of parameters to balance the different structure complexities of them as show in Table \ref{t:parameters}. 

\begin{table}[h]
\begin{center}
\begin{tabular}{|l|l|l|l|l|}
\hline \multicolumn{4}{|c|}{sfIN} & biLSTM-crf\\
\hline Component & \textbf{TE} & {SC} & Total & Total\\
\hline parameters & 49876 & 167936 & 217812 & 196608\\
\hline
\end{tabular}
\end{center}
\caption{Parameters Numbers}
\label{t:parameters}
\end{table}

%\begin{itemize}
%\item \textsl{sfIN-bl}: sfIN with both supervised and reinforcement learning 
%\item \textsl{sfIN-srl}: sfIN with large $\lambda$ value to enhance reinforcement learning
%\end{itemize}

In both incident identification task on \textsl{TcJD} and focuses of dispute identification task on \textsl{PiJD}, we randomly sample 500 documents as the test set and the other documents are used in training procedure.

\section{Results and Discussion} 
In this section, we are going to show results of experiments and explain some interesting phenomena during the test procedure. 
\begin{table*}[h] \footnotesize
	\begin{center}
		\begin{tabular}{|l|l|l|l|l|l|}
			\toprule
			model & WA(\%) & precision(\%) & recall(\%) & F1(\%) & $wlar$  \\
			\hline \multicolumn{6}{|c|}{TaskI} \\
			\toprule
			biLSTM-crf & $97.29$ & $78.62$ & $85.26$ & $81.81$ & 1 \\
			ZN  & $99.93$ & $95.34$ & $95.64$ & $95.49$ & 0.16 \\
			\hline \multicolumn{6}{|c|}{TaskII} \\
			\toprule
			biLSTM-crf & $99.61$ & $78.31$ & $86.83$ & $82.36$ & 1\\
			ZN & $99.92$ & $94.74$ & $95.06$ & $94.91$ & 0.15\\
			\hline
		\end{tabular}
	\end{center}
	\caption{Evaluation Results, WA: wrod-level accuracy}
	\label{t:result}
\end{table*}

Table \ref{t:result} presents our comparisons of \texttt{biLSTM-crf} and \ZN for \textbf{TaskI}(incident identification) and \textbf{TaskII}(focuses identification). \ZN outperforms \texttt{biLSTM-crf} by more than 10 F1-measure on both tasks.
Though \texttt{biLSTM-crf} and other similar sequence labeling models achieve great successes on several tasks, they fail to perform well on long text for several reasons: (1) Although an LSTM has a self-loop for the gradients that helps to capture the long-term dependencies by mitigating the vanishing gradient problem, in practice, it is still limited to a few hundred time steps due to the leaky integration by which the contents to memorize for a long-term is gradually diluted at every time step.\cite{chung2016hierarchical} (2) Dependencies in such long texts simultaneously involve multi-level information, which is very hard for models not capable of explicitly leveraging text structures.

In contrast, \ZN has following advantages: (1) With text structure hierarchies, representation directly corresponds to language structures. This allows \ZN to precisely integrate multi-level information in modeling complex long-term dependencies. (2) Taking high-level actions significantly reduces the time steps between decision and the information it needs, thus it provides short-cuts for gradient back-propagation. (3) \ZN is more efficiency in processing. As shown in Table \ref{t:result}, only 15\% of actions \ZN used are at word-level.

\begin{figure}[h]
	\centering
	\includegraphics[width=0.4\textwidth]{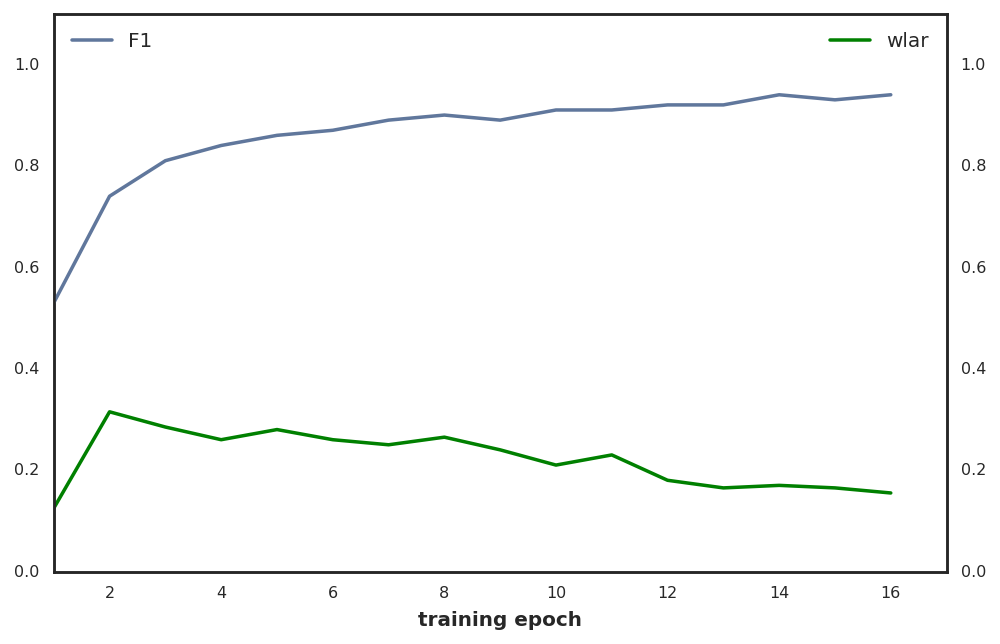}
	\caption{average $wlar$ and $F1$ of validation set during training}
	\label{fig:curve}
\end{figure}

Figure \ref{fig:curve} shows average $wlar$ and $F1$ of validation set during a simple \ZN training procedure. By jointly using supervised learning and reinforcement learning, the model learns to simultaneously improve accuracy and efficiency.

\begin{figure}[h]
	\centering
	\includegraphics[width=0.45\textwidth]{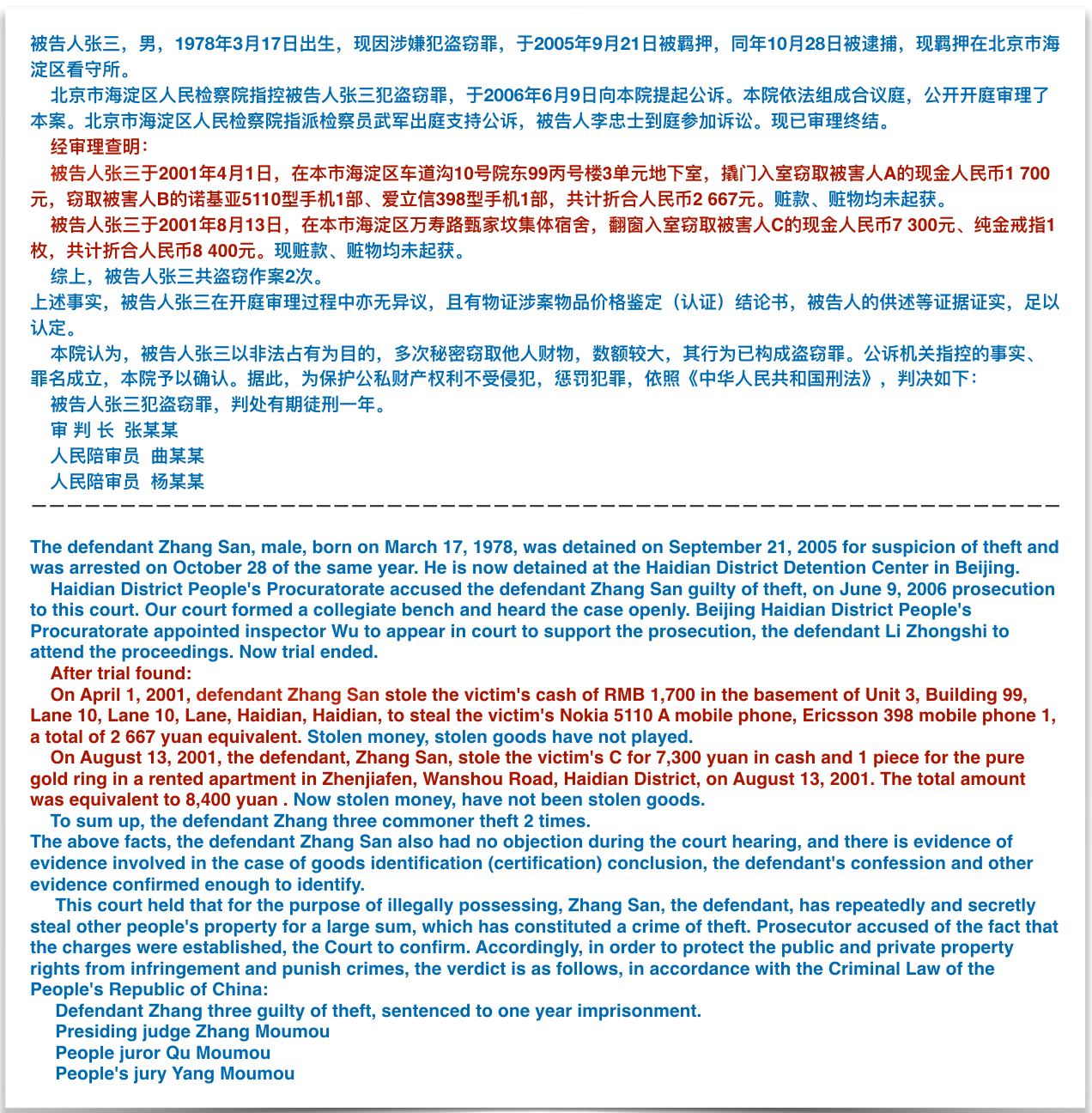}
	\caption{processing procedure on a fragment of a simple case: red part indicates using word-level actions and blue part indicates using sentence-level or paragraph-level actions}
	\label{fig:case}
\end{figure}

To gain a more intuitive understanding how \ZN works, we randomly choose a sample from \textsl{TcJD} dataset, as shown in Figure \ref{fig:case}. The processing path \ZN takes is very intuitive: (1) Irrelevant sections such as \textsl{personal information} and the \textsl{judgment} are processed using high-level actions. (2) Fragments that obviously indicating clue for target information and target information itself are processed by word-level actions. This strategy is exactly what humans use to choose between coarse reading or intensive reading.
%\section{Results and Discussion} 
%\subsection{Basic Result}
%\subsection{Path Analysis}
\section{Conclusion}
The proposed model \ZN is capable of building an effective multi-resolution representation and developing its own analyzing rhythm by choosing different actions combinations to extract critical information. Trained with a hybrid paradigm of supervised learning (distinguishing right and wrong decision) and reinforcement learning (determining the goodness among multiple right paths), our model has both high accuracy and efficiency. The experiments show that the model achieves higher performance than the state-of-the-art sequence labeling model in long text labeling tasks. 
\newpage



\bibliographystyle{elsarticle-num} 
\bibliography{ZN.bib}


%\begin{thebibliography}{00}


%\end{thebibliography}
\end{document}